\pdfoutput=1

\documentclass[11pt]{article}
\usepackage{float}
\usepackage[final]{acl}
\usepackage{times}
\usepackage{latexsym}
\usepackage{subcaption}
\usepackage{multirow} 
\usepackage{adjustbox}
\usepackage{tabularx}
\usepackage{booktabs}
\usepackage{makecell}
\usepackage{ragged2e}

\usepackage{enumitem}
\usepackage{caption}
\usepackage{adjustbox}
\usepackage{array}
\newcolumntype{P}[1]{>{\centering\arraybackslash}p{#1}}
\newcolumntype{M}[1]{>{\centering\arraybackslash}m{#1}}

\usepackage[T1]{fontenc}

\usepackage[utf8]{inputenc}

\usepackage{microtype}

\usepackage{inconsolata}

\usepackage{graphicx}

\title{Cyber for AI at SemEval-2025 Task 4: Forgotten but Not Lost: The Balancing Act of Selective Unlearning in Large Language Models}

\author{Dinesh Srivasthav P \\
  TCS Research \\
  \texttt{dineshsrivasthav.p@tcs.com} \\\And
   Bala Mallikarjunarao Garlapati \\
  TCS Research \\
  \texttt{balamallikarjuna.g@tcs.com}} 

\begin{document}
\maketitle
\begin{abstract}
Large Language Models (LLMs) face significant challenges in maintaining privacy, ethics, and compliance, when sensitive or obsolete data must be selectively removed. Retraining these models from scratch is computationally infeasible, necessitating efficient alternatives. As part of the SemEval 2025 Task 4, this work focuses on the application of selective unlearning in LLMs to address this challenge. In this paper, we present our experiments and findings, primarily leveraging global weight modification to achieve an equilibrium between effectiveness of unlearning, knowledge retention, and target model's post-unlearning utility. We also detail the task-specific evaluation mechanism, results, and challenges. Our algorithms have achieved an aggregate score of 0.409 and 0.389 on the test set for 7B and 1B target models, respectively, demonstrating promising results in verifiable LLM unlearning. 
\end{abstract}

\section{Introduction}
Large Language Models (LLMs) have revolutionized the way artificial intelligence can be used, adapted, and integrated, demonstrating unprecedented capabilities across various domains and use cases \cite{NEURIPS2020_1457c0d6}. In order for LLMs to provide optimal and factual responses, they often require to be trained on vast amount of diverse information which not only includes the world data, but could also contain sensitive application-, task-, or entity-specific data \cite{10.1145/3442188.3445922}. Training on such massive datasets typically introduces critical challenges related to bias, ethics, and privacy concerns \cite{neel2024privacyissueslargelanguage, ZHANG2025100301}. Further, at times, the data providers might also want to have no traces of their data at a later point due to reasons such as confidentiality, legal issues, change in terms, etc \cite{yao2024largelanguagemodelunlearning}. Retraining LLMs to exclude specific data is computationally expensive and impractical \cite{cottier2025risingcoststrainingfrontier, xia2024understandingperformanceestimatingcost}, especially given the potential for numerous subsequent removal requests from various data providers, clients, or end-users. 

\begin{figure}[t]
  \includegraphics[width=\columnwidth]{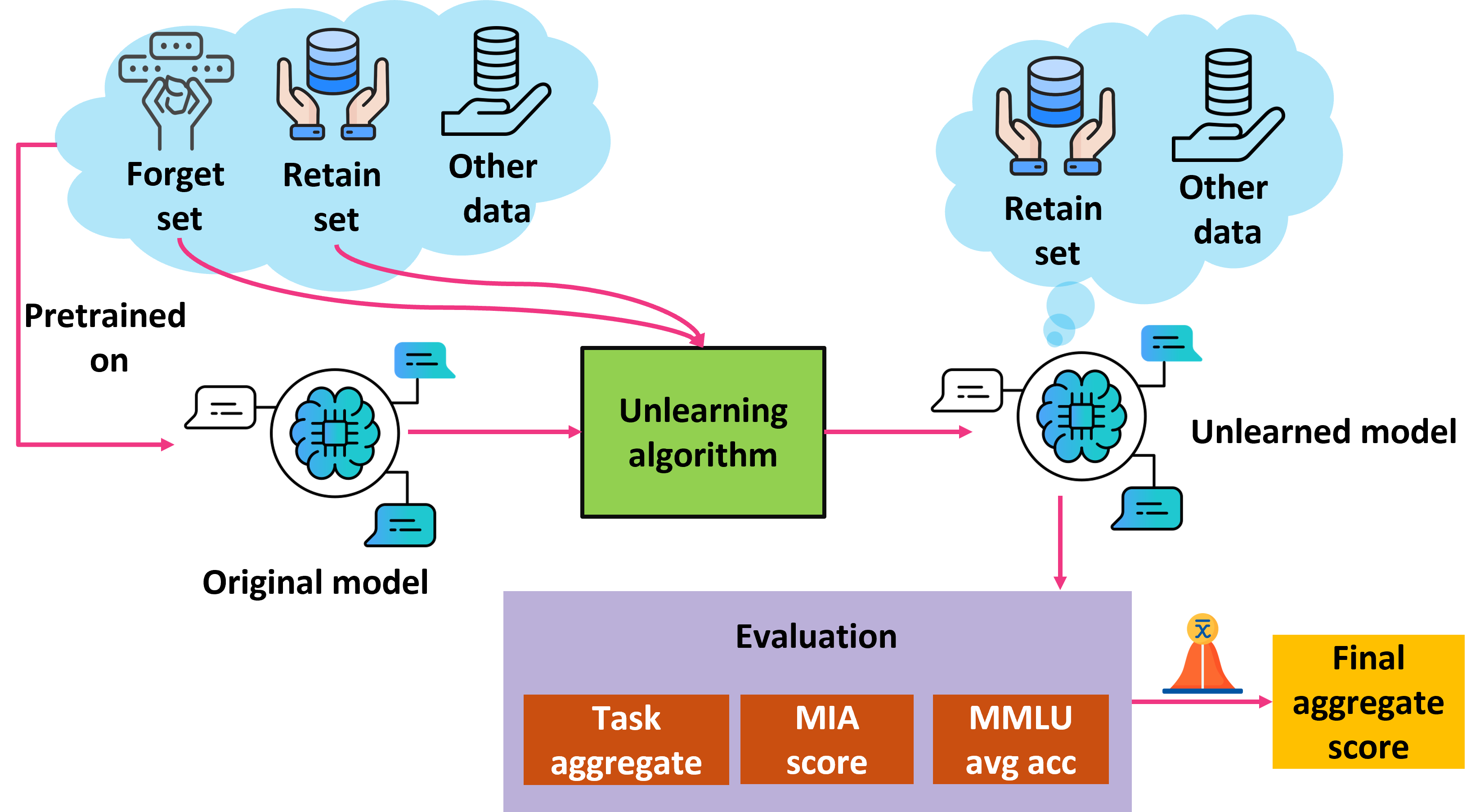}
  \caption{Block diagram of selective unlearning in LLMs, with task's evaluation mechanism}
  \label{fig:bd}
  \vspace{-5mm}
\end{figure}
Selective unlearning in LLMs \cite{liu2024rethinkingmachineunlearninglarge} helps to achieve this exact objective. It is a mechanism through which the requested information can be precisely removed from the model's parametric memory along with preserving the model's knowledge integrity and utility for downstream tasks, without retraining it from scratch. The requested information can be a specific knowledge, model's certain behavior, a feature, or its ability to perform a particular task, or a combination of two or more of these and more. Figure \ref{fig:bd} depicts the crux of selective unlearning along with the task's evaluation mechanism which is discussed in Section \ref{sec:evaluation}.\\ 
Unlearning in LLMs is a niche yet increasingly critical area that offers key advantages such as cost-effectiveness, computational efficiency, and precise intervention. It plays a crucial role in eliminating embedded biases \cite{yu-etal-2023-unlearning, dige-etal-2024-machine}, erasing toxic or harmful responses \cite{liu-etal-2024-towards-safer}, and reinforcing AI guardrails \cite{Hine2024} in safety-critical fields such as healthcare, finance, and enterprise settings. However, ensuring effective and verifiable unlearning is challenging, as many approaches risk leaving residual traces of removed knowledge or inadvertently impairing broader model capabilities. Achieving the right balance between unlearning effectiveness, knowledge retention, and model generalization is a delicate optimization problem that continues to drive research in this field \cite{qu2024frontierdataerasuremachine}.

\section{Related works}
The approaches to unlearning in LLMs can be broadly classified into four categories: global weight modification, local weight modification, architecture modification, input/output modification \cite{BlancoJusticia2025}.\\ 
Global weight modification involves updating all the model parameters while unlearning, thus, ensuring better guarantee of forgetting the requested information. It includes approaches such as gradient ascent \cite{feng-etal-2024-fine, gundavarapu2024machineunlearninglargelanguage}, gradient difference \cite{bu2024unlearningmultitaskoptimizationnormalized}, knowledge distillation \cite{zhao2024unlearningbackdoorattacksllms}, KL minimization \cite{yao2024largelanguagemodelunlearning}, weight perturbation \cite{yuan2024robustknowledgeunlearningadversarial}, and so on. These approaches are well suited for smaller models and provide strong unlearning, however, are resource intensive for larger models, as the training costs greatly increase with increase in the number of parameters. Global weight modification for larger models also strengthens the problem of optimizing effective unlearning, and preserving model's capabilities.\\ 
Local weight modification identifies a subset of parameters that are required to be modified and accordingly updates only those model parameters \cite{ashuach2024revsunlearningsensitiveinformation, wu-etal-2023-depn, jia2024waglestrategicweightattribution, pochinkov2024dissectinglanguagemodelsmachine}, thereby, minimizing the computational efforts needed. Nevertheless, the right set of parameters that are required to be modified might vary based on the diversity of the requested information. Identifying the same is thus, challenging which therefore, has chances of leaving traces of unlearning, or in other words, influence of the requested information could still be observed in the model's behavior \cite{hong2024intrinsicevaluationunlearningusing}.\\ 
Architecture modification based approaches involve tweaking the model's architecture such as by adding additional layers \cite{chen-yang-2023-unlearn}, or by using external modules \cite{ji2024reversingforgetretainobjectivesefficient, zhang2023composingparameterefficientmodulesarithmetic} in addition to the target model, etc. These approaches, while advantageous in other contexts, were not suitable for this specific task's setup.\\
Finally, the input/output modification, as the name suggests, involves approaches that do not achieve true unlearning but modifies the model's input, and sometimes the output, in such a way that the final response is as desired, by leveraging techniques such as soft prompting \cite{bhaila2024softpromptingunlearninglarge}, preference optimization \cite{zhang2024negative, fan2024simplicityprevailsrethinkingnegative}, in-context learning \cite{pawelczyk2023context, thaker2024guardrailbaselinesunlearningllms}.

\section{Task artifacts}
This section describes the artifacts given by the task organizers namely: dataset, models, and MIA dataset, which have been used for this task of unlearning.

\begin{table}
    \centering
    \footnotesize
    \begin{tabular}{lcc}
        \toprule
        \textbf{Split} & \textbf{Train} & \textbf{Validation} \\
        \midrule
        Forget & 1112 & 254 \\
        Retain & 1136 & 278 \\
        \bottomrule
    \end{tabular}
    \caption{Train and validation splits of the datasets}
    \label{tab:data_splits}
\end{table}

\subsection{Dataset}
There are two disjoint components of the dataset: forget dataset and retain dataset. As their names suggest, forget dataset constitutes of samples that have to be forgotten or unlearned by the model, and the retain dataset constitutes of samples that still have to be retained by the model post its unlearning.
The dataset has predefined splits between train and validation sets. The sample distribution between the train and validation sets of forget and retain sets is respectively presented in Table \ref{tab:data_splits}. 
Each of the forget and retain datasets in their json format have five fields as described in Table \ref{tab:fields_description}. Further as described in Table \ref{tab:fields_description}, there are three tasks to which a sample in forget dataset and retain dataset could belong to as follows:
    (1) Long-form synthetic creative documents across genres;
    (2) Short-form synthetic biographies with PII (fake names, phone numbers, SSNs, emails, addresses);
    (3) Real documents sampled from the target model’s training dataset.\\
The forget and retain data samples were designed to be evaluated on sentence completion, and question-answering. Therefore, the input field in the retain and forget datasets is either an excerpt from some document, or is a question. While the output field is the continuation of the corresponding input if the input is a document excerpt (sentence completion), or is an answer if the input is a question (question-answering). This categorization is indicated with a string -- 'sc' or 'qa' as part of the sample's id field. \cite{ramakrishna2025lumellmunlearningmultitask} further discusses the process of dataset curation.

\subsection{Models}
Two models were given as the target models that need to unlearn the forget dataset. One is a 7-billion parameter model, and the other is a 1-billion model, both finetuned to memorize the forget and the retain datasets, with their base architectures being OLMo-7B-0724-Instruct-hf\footnote{\url{https://huggingface.co/allenai/OLMo-7B-0724-Instruct-hf}} model and OLMo-1B-0724-hf\footnote{\url{https://huggingface.co/allenai/OLMo-1B-0724-hf}} model respectively.\\ 
The base models OLMo-7B-0724-Instruct-hf and OLMo-1B-0724-hf are transformer style autoregressive language models from the family of Open language Models (OLMo) by Allen Institute for AI, and were trained on Dolma dataset\footnote{\url{https://huggingface.co/datasets/allenai/dolma}}, with the Instruct version trained on UltraFeedback dataset\footnote{\url{https://huggingface.co/datasets/allenai/ultrafeedback_binarized_cleaned}}. Dolma is a large dataset curated from a combination of diverse materials sourced from the internet, academic journals, published literature, software repositories, books, and so on. The UltraFeedback dataset is a large collection of human feedback including human preferences and ratings for different LLM outputs. 

\begin{table}[t]
  \centering
  \footnotesize
  \begin{tabular}{p{1cm}|p{6cm}}
    \toprule
    \textbf{Field} & \textbf{Description} \\
    \midrule
    id     & Document id           \\
    input     & Document snippet (input to the model)          \\
    output     & Output for the corresponding input based on the concerned task mapped           \\
    task     & The respective unlearning task to which the sample is assigned from the three tasks            \\
    split     & If the sample belongs to retain or forget set           \\ \bottomrule
  \end{tabular}
  \caption{Field description of forget and retain datasets}
  \label{tab:fields_description}
\end{table}

\subsection{Evaluation}
\label{sec:evaluation}
The target model's unlearning is evaluated as an average of three different scores namely task aggregate, MIA score, and MMLU average accuracy, which are explained as follows.\\
    \textbf{Task aggregate}: 
    All the samples in the forget and retain datasets are respectively grouped according to one-of-the-three task mapping. For all the samples in these six sets, Regurgitation score is computed as RougeL score for samples in sentence completion format, and Knowledge score is a binary indicator computed as the exact match rate for the samples in question-answer format. The scores are inverted ($1-score$) if the sample is part of the forget dataset. The respective scores for each of the six sets are aggregated and a harmonic mean of these 12 scores is considered as the task aggregate. A higher score represents better performance.\\
    \textbf{MIA score}: Membership Inference Attack (MIA) is typically used to know if a sample is part of trained model's training data or not. In the context of unlearning, it is therefore, used to identify whether the samples in the forget dataset were forgotten by the model or not.\\ 
    The task organizers have given an MIA dataset for this purpose which constitutes of two sets: Member set and Non-member set, each with 150 samples. Member set is a subset of the train split of the forget dataset. Non-member set constitutes of samples collected from elsewhere which the model has not seen prior. 
    Both the member and non-member sets are given in jsonl format. Each sample in the member set has an id field, a document field, a question\_answering\_task field constituting of a question and the corresponding answer, a sentence\_completion\_task field constituting of an input and the corresponding output.
    Each sample in the non-member set has certain meta fields, and a document excerpt.\\
    The final MIA score is computed as $1-abs(mia\_auc) - 0.5)*2$ where $mia\_auc$ is the area under the receiver operating characteristic curve with the negative log likelihoods computed for member and non-member sets. The MIA score is expected to be around 0.5. The closer it is to zero denotes under-unlearning, and the closer it is to one denotes over-unlearning.\\
    \textbf{MMLU average accuracy}: MMLU stands for Massive Multitask Language Understanding \cite{hendrycks2021measuringmassivemultitasklanguage}. It is a benchmark dataset consisting of 15908 multiple-choice questions spanning across 57 diverse subjects used for evaluating various capabilities of language models such as language understanding, general knowledge, reasoning abilities, domain knowledge, generalization, and so on. The average accuracy of the target model across all the 57 subjects is used as one of the metrics to evaluate the post-unlearning utility of the target model. A higher score represents better utility. A threshold of 0.371 is set by the organizers.

\section{Experiments and Results}

A variety of experiments have been tried to understand the patterns in the given datasets, and figure out the suitable approaches that would balance the tradeoff between target model's unlearning with its utility post unlearning. Some of them have been discussed in Appendix \ref{sec:B}. We used a Nvidia GeForce RTX A6000 (48GB) GPU to run the experiments.

\begin{table}[t]
\centering\footnotesize
\setlength{\tabcolsep}{1.5pt} 
\begin{tabular}{p{2.8cm}p{1.5cm}p{0.84cm}p{1cm}p{1.1cm}}
\toprule
\textbf{Method}     & \textbf{Aggregate} & \textbf{Task Agg.} & \textbf{MIA score} & \textbf{MMLU Avg Acc.} \\ \midrule
Gradient Ascent     & 0.345              & 0                       & 0.807              & 0.229                 \\ 
Controlled GA       & 0.370$^\circ$              & 0                       & 0.855              & 0.255                 \\ 
Gradient Difference & 0.360$^*$              & 0                       & 0.825              & 0.255                 \\ 
KL Minimization     & 0.174              & 0.219                   & 0.032              & 0.272                 \\ 
Xavier init (1B)    & 0.402$^\circ$             & 0                       & 0.944              & 0.261                 \\ 
Original model (1B) & 0.0913             & 0                       & 0                  & 0.274                 \\ 
Gradient descent    & 0.410$^\circ$             & 0                       & 0.982              & 0.247                 \\ 
Test set score      & 0.389             & 0                       & 0.914              & 0.251                 \\ \bottomrule
\end{tabular}
\captionsetup{justification=centering}
\caption{Performance of 1B model\\(Higher than submission$^\circ$ ; Submission$^*$)}
\label{1B_short_table}
\end{table}

\begin{table}[h!]
\centering
\footnotesize
\setlength{\tabcolsep}{1.5pt} 
\begin{tabular}{M{2.6cm}M{1.6cm}M{0.84cm}M{1cm}M{1.2cm}}
\toprule
\textbf{Method} & \textbf{Aggregate} & \textbf{Task Agg.} & \textbf{MIA score} & \textbf{MMLU Avg Acc.} \\ \midrule \vspace{0.1cm}
Gradient Ascent & 0.383 & \phantom{0}0\phantom{.005} & \phantom{0}0.865 & \phantom{0}0.284 \\ \vspace{0.1cm}
Gradient Difference & 0.171 & \phantom{0}0\phantom{.005} & \phantom{0}0\phantom{.865} & \phantom{0}0.512 \\ \vspace{0.1cm}
Gradient Difference -> Gradient Ascent & 
    \begin{tabular}[c]{@{}l@{}}\phantom{0}0.377$^*$\\ \phantom{0}0.447$^*$\end{tabular} & 
    \begin{tabular}[c]{@{}l@{}}\phantom{0}0\phantom{.005}\\ \phantom{0}0\phantom{.005}\end{tabular} & 
    \begin{tabular}[c]{@{}l@{}}\phantom{0}0.670\\ \phantom{0}0.998\end{tabular} & 
    \begin{tabular}[c]{@{}l@{}}\phantom{0}0.461\\ \phantom{0}0.343\end{tabular} \\ \vspace{0.1cm}
Gradient Difference -> Gradient Difference & \phantom{0}0.171$^*$ & \phantom{0}0\phantom{.005} & \phantom{0}0\phantom{.936} & \phantom{0}0.513 \\ \vspace{0.1cm}
Xavier init (7B) & 0.397 & \phantom{0}0\phantom{.005} & \phantom{0}0.936 & \phantom{0}0.255 \\ \vspace{0.1cm}
Original model (7B) & 0.170 & \phantom{0}0\phantom{.005} & \phantom{0}0\phantom{.936} & \phantom{0}0.512 \\ \vspace{0.1cm}
Gradient Descent & 0.170 & \phantom{0}0.005 & \phantom{0}0\phantom{.847} & \phantom{0}0.504 \\ \vspace{0.1cm}
Gradient Descent -> Gradient Ascent & 0.365 & \phantom{0}0\phantom{.005} & \phantom{0}0.847 & \phantom{0}0.247 \\\vspace{0.1cm}
Test set score & 0.409 & \phantom{0}0\phantom{.005} & \phantom{0}0.999 & \phantom{0}0.229 \\ \bottomrule
\end{tabular}
\caption{Performance of 7B model (Submissions$^*$) }
\label{7B_short_table}
\vspace{-4.5mm}
\end{table}
\hspace{-10cm}

\noindent \textbf{Gradient-based methods:}\\
Due to computational constraints, we have considered two configurations of the models for executing these methods: 1B model is trained as is, and 7B model is trained in a 4-bit PEFT configuration, described in Table \ref{training_config_table} of Appendix \ref{sec:A}. Due to this, some of the experiments have been performed on either of these models, and some of them on both the models. The study investigates how various gradient modifications influence performance, with a focus on balancing retention, unlearning, and model utility. The key results of the same are briefly reported in Tables \ref{1B_short_table}, \ref{7B_short_table} for 1B and 7B models respectively. 
A detailed comparison of all the variants of these experiments along with the corresponding training configurations is presented in Tables \ref{1B_full_table}, \ref{7B_full_table} of Appendix \ref{sec:A} for 1B and 7B models respectively.\\
    \noindent \textbf{Gradient ascent:}
    In this method, the target model was trained on the forget set with an inverted loss, thereby, making it to unlearning the training set rather than learning it. By maximizing the loss on the forget set, this method forces the model to unlearn. It was observed that learning rate (LR) and weight decay (WD) variations have a strong impact on the unlearning intensity. In particular, aggressive configurations led to stronger unlearning, achieving a MIA score of 0.807 by the 1B model, when LR and WD were increased despite halving the number of epochs (E). On the other hand, with a steady training for 10 epochs, with relatively lesser LR, achieved even more unlearning in the 7B model, however, costing its utility -- the model's MMLU average accuracy (MMLUAA) almost got halved compared to the original 7B model. Nonetheless, it was noted that the MMLUAA has not dropped much in the case of 1B model, despite reaching a similar MIA score. This indicates that gradient ascent though achieves a good balance between model's utility and the level of unlearning in smaller models, it tends to easily destabilize large models. Quantization in the 7B model may have also enhanced unlearning, potentially due to increased numerical instability aiding divergence from the learned state.

    \noindent \textbf{Gradient descent:}
    In this method, the target model was trained on the retain set, thereby, optimizing it to the training set. The intuition is that, the strong adaption of the model only to the retain set can naturally make it tend to forget the other information (forget set) it was previously trained on, like catastrophic forgetting. A few interesting observations were made by the model performances with this method. Firstly, with 1B model, when it was trained for 6 epochs, it has reached the optimal level of unlearning required ($\mathord{\sim}$ 0.5), and also got a slight boost in the MMLUAA, even more than the original 1B model by 0.001. However, the overall score (aggregate) is not high. It is even significantly less than one of the gradient ascent scores. To further study if the model's performance would be increased if trained more aggressively, the epochs were increased to 20, besides increasing LR and WD. This has achieved near perfect unlearning, and also is the highest MIA score amongst all the experiments on the 1B model. The MMLUAA has also not dropped much from the original 1B model, and therefore has got the highest aggregate score of 0.410 on the 1B model. However, a similar setup did not go well with the 7B model. 
    
    \noindent \textbf{KL minimization:} This method adds an additional term of Kullback–Leibler (KL) divergence as regularization to the loss function. This was used in gradient ascent with the objective of maximizing the loss on the forget set, yet not deviate too drastically from the original model, preserving its performance. However, we observed that KL minimization was not much different from the gradient ascent when executed with same parameters.
    
    \noindent \textbf{Controlled gradient ascent:} We tried with a variant of gradient ascent where gradients were modified in a controlled manner instead of completely getting updated based on change with loss. A parameter $alpha$ was used to control the scale of updation. Setting alpha to 0.1 helped regulate the magnitude of updates, allowing the 1B model to reach a MIA score of 0.855, outperforming standard gradient ascent, despite training aggressively for more than triple the epochs. This approach was particularly effective in preventing the complete collapse of model utility, making it a viable strategy when unlearning must be balanced against task performance.
    
    \noindent \textbf{Gradient difference:} To reinforce the model utility degraded by gradient ascent, in this method, the target model was further trained on the retain set, thereby, optimizing it to the training set. Therefore, the target model goes through gradient ascent followed by gradient descent. While this method has shown a steady increase in the aggregate score of the 1B model, with a tradeoff between MIA score and MMLUAA, it has not shown any impact on the 7B model. It is important to note that the 10 epochs of gradient ascent has brought down the MMLUAA to 0.284 from 0.512, and a single subsequent gradient descent epoch with LR as low as 2e-6 brought it back to 0.511, emphasizing the vital role and impact of gradient descent in mitigating utility loss in larger models.\\
    \noindent \textbf{Gradient difference followed by Gradient ascent:} To further study the behavior of 7B model, given its drastic and static responses to the above methods, we made a few experiments specifically on the 7B model such as this, and the subsequent ones. It was observed that one epoch of gradient ascent with a slightly higher LR has shown drastic change in model's performance. Multiple experiments with this method demonstrate its strategic impact striking a good balance between unlearning and utility. Nevertheless, reducing the learning rate has not reversed the impact of gradient descent, while increasing it further has deteriorated the model's performance with excessive unlearning like that of gradient ascent alone. Overall, with optimal parameter settings, this method excelled on the 7B model, achieving the highest aggregate score while maintaining a balance across utility and unlearning metrics.\\
    \noindent \textbf{Gradient difference followed by Gradient difference:} Further gradient descent on the aforementioned state reemphasizes the impact of even a single round of gradient descent with LR as low as 2e-8, on a strongly unlearned model with MIA score of 0.982 which reinstated it alike the original. \\ 
    \noindent \textbf{Gradient descent followed by Gradient ascent:} It was observed that gradient ascent with smaller learning rate like 2e-6 could not counter the impact of prior gradient descent training, while making it a little aggressive has over dominated the prior training, leading to reduced MMLUAA.\\    
    \noindent \textbf{Xavier Initialization:} Though, not an unlearning method originally, it is interesting to observe that by erasing all the parametric values of the original models and by only initializing them with Xavier initialization has still given one of the best aggregate scores, without any training, outperforming many other methods, stressing setup-dependent variability.  
    

\section{Conclusion}
This work explores the use of targeted unlearning in LLMs where we have experimented with several unlearning methods with different configurational settings to make the target models forget the requested dataset, and preserve the specified retain set, also, preserving its overall multifaceted capabilities. From our experiments,
for 7B model: Gradient difference followed by Gradient ascent worked well with appropriate parameters tuning; and
for 1B model: Gradient descent alone on the retain set worked well. Xavier initialization on the 1B model has got near equivalent score on all the metrics as the former. Followed by similar performance between Controlled gradient ascent, and Gradient difference, with respective appropriate parameters tuning. 
This work reemphasizes the fact that selective unlearning comes with the delicate problem of optimizing effective unlearning with knowledge retention of the remaining data and model's integrity, utility for downstream tasks. Further, it demonstrates that performance of a method significantly depends on the scale of the target model, and the kind of data it is presented with.

\section*{Acknowledgments}

We would like to thank our colleagues, Ashok Urlana and Charaka Vinayak Kumar, for their valuable feedback and insightful discussions, which have notably improved this work.

\bibliography{acl_latex}

\begin{thebibliography}{34}
\providecommand{\natexlab}[1]{#1}

\bibitem[{Ashuach et~al.(2024)Ashuach, Tutek, and Belinkov}]{ashuach2024revsunlearningsensitiveinformation}
Tomer Ashuach, Martin Tutek, and Yonatan Belinkov. 2024.
\newblock \href {https://arxiv.org/abs/2406.09325} {Revs: Unlearning sensitive information in language models via rank editing in the vocabulary space}.
\newblock \emph{Preprint}, arXiv:2406.09325.

\bibitem[{Bender et~al.(2021)Bender, Gebru, McMillan-Major, and Shmitchell}]{10.1145/3442188.3445922}
Emily~M. Bender, Timnit Gebru, Angelina McMillan-Major, and Shmargaret Shmitchell. 2021.
\newblock \href {https://doi.org/10.1145/3442188.3445922} {On the dangers of stochastic parrots: Can language models be too big?}
\newblock In \emph{Proceedings of the 2021 ACM Conference on Fairness, Accountability, and Transparency}, FAccT '21, page 610–623, New York, NY, USA. Association for Computing Machinery.

\bibitem[{Bhaila et~al.(2024)Bhaila, Van, and Wu}]{bhaila2024softpromptingunlearninglarge}
Karuna Bhaila, Minh-Hao Van, and Xintao Wu. 2024.
\newblock \href {https://arxiv.org/abs/2406.12038} {Soft prompting for unlearning in large language models}.
\newblock \emph{Preprint}, arXiv:2406.12038.

\bibitem[{Blanco-Justicia et~al.(2025)Blanco-Justicia, Jebreel, Manzanares-Salor, Sánchez, Domingo-Ferrer, Collell, and Eeik~Tan}]{BlancoJusticia2025}
Alberto Blanco-Justicia, Najeeb Jebreel, Benet Manzanares-Salor, David Sánchez, Josep Domingo-Ferrer, Guillem Collell, and Kuan Eeik~Tan. 2025.
\newblock \href {https://doi.org/10.1007/s10462-024-11078-6} {Digital forgetting in large language models: a survey of unlearning methods}.
\newblock \emph{Artificial Intelligence Review}, 58(3).

\bibitem[{Brown et~al.(2020)Brown, Mann, Ryder, Subbiah, Kaplan, Dhariwal, Neelakantan, Shyam, Sastry, Askell, Agarwal, Herbert-Voss, Krueger, Henighan, Child, Ramesh, Ziegler, Wu, Winter, Hesse, Chen, Sigler, Litwin, Gray, Chess, Clark, Berner, McCandlish, Radford, Sutskever, and Amodei}]{NEURIPS2020_1457c0d6}
Tom Brown, Benjamin Mann, Nick Ryder, Melanie Subbiah, Jared~D Kaplan, Prafulla Dhariwal, Arvind Neelakantan, Pranav Shyam, Girish Sastry, Amanda Askell, Sandhini Agarwal, Ariel Herbert-Voss, Gretchen Krueger, Tom Henighan, Rewon Child, Aditya Ramesh, Daniel Ziegler, Jeffrey Wu, Clemens Winter, Chris Hesse, Mark Chen, Eric Sigler, Mateusz Litwin, Scott Gray, Benjamin Chess, Jack Clark, Christopher Berner, Sam McCandlish, Alec Radford, Ilya Sutskever, and Dario Amodei. 2020.
\newblock \href {https://proceedings.neurips.cc/paper_files/paper/2020/file/1457c0d6bfcb4967418bfb8ac142f64a-Paper.pdf} {Language models are few-shot learners}.
\newblock In \emph{Advances in Neural Information Processing Systems}, volume~33, pages 1877--1901. Curran Associates, Inc.

\bibitem[{Bu et~al.(2024)Bu, Jin, Vinzamuri, Ramakrishna, Chang, Cevher, and Hong}]{bu2024unlearningmultitaskoptimizationnormalized}
Zhiqi Bu, Xiaomeng Jin, Bhanukiran Vinzamuri, Anil Ramakrishna, Kai-Wei Chang, Volkan Cevher, and Mingyi Hong. 2024.
\newblock \href {https://arxiv.org/abs/2410.22086} {Unlearning as multi-task optimization: A normalized gradient difference approach with an adaptive learning rate}.
\newblock \emph{Preprint}, arXiv:2410.22086.

\bibitem[{Chen and Yang(2023)}]{chen-yang-2023-unlearn}
Jiaao Chen and Diyi Yang. 2023.
\newblock \href {https://doi.org/10.18653/v1/2023.emnlp-main.738} {Unlearn what you want to forget: Efficient unlearning for {LLM}s}.
\newblock In \emph{Proceedings of the 2023 Conference on Empirical Methods in Natural Language Processing}, pages 12041--12052, Singapore. Association for Computational Linguistics.

\bibitem[{Cottier et~al.(2025)Cottier, Rahman, Fattorini, Maslej, Besiroglu, and Owen}]{cottier2025risingcoststrainingfrontier}
Ben Cottier, Robi Rahman, Loredana Fattorini, Nestor Maslej, Tamay Besiroglu, and David Owen. 2025.
\newblock \href {https://arxiv.org/abs/2405.21015} {The rising costs of training frontier ai models}.
\newblock \emph{Preprint}, arXiv:2405.21015.

\bibitem[{Dige et~al.(2024)Dige, Arneja, Yau, Zhang, Bolandraftar, Zhu, and Khattak}]{dige-etal-2024-machine}
Omkar Dige, Diljot Arneja, Tsz~Fung Yau, Qixuan Zhang, Mohammad Bolandraftar, Xiaodan Zhu, and Faiza~Khan Khattak. 2024.
\newblock \href {https://doi.org/10.18653/v1/2024.emnlp-industry.71} {Can machine unlearning reduce social bias in language models?}
\newblock In \emph{Proceedings of the 2024 Conference on Empirical Methods in Natural Language Processing: Industry Track}, pages 954--969, Miami, Florida, US. Association for Computational Linguistics.

\bibitem[{Fan et~al.(2024)Fan, Liu, Lin, Jia, Zhang, Mei, and Liu}]{fan2024simplicityprevailsrethinkingnegative}
Chongyu Fan, Jiancheng Liu, Licong Lin, Jinghan Jia, Ruiqi Zhang, Song Mei, and Sijia Liu. 2024.
\newblock \href {https://arxiv.org/abs/2410.07163} {Simplicity prevails: Rethinking negative preference optimization for llm unlearning}.
\newblock \emph{Preprint}, arXiv:2410.07163.

\bibitem[{Feng et~al.(2024)Feng, Chen, Li, and Lin}]{feng-etal-2024-fine}
XiaoHua Feng, Chaochao Chen, Yuyuan Li, and Zibin Lin. 2024.
\newblock \href {https://doi.org/10.18653/v1/2024.emnlp-main.566} {Fine-grained pluggable gradient ascent for knowledge unlearning in language models}.
\newblock In \emph{Proceedings of the 2024 Conference on Empirical Methods in Natural Language Processing}, pages 10141--10155, Miami, Florida, USA. Association for Computational Linguistics.

\bibitem[{Gundavarapu et~al.(2024)Gundavarapu, Agarwal, Arora, and Jagadeeshaiah}]{gundavarapu2024machineunlearninglargelanguage}
Saaketh~Koundinya Gundavarapu, Shreya Agarwal, Arushi Arora, and Chandana~Thimmalapura Jagadeeshaiah. 2024.
\newblock \href {https://arxiv.org/abs/2405.15152} {Machine unlearning in large language models}.
\newblock \emph{Preprint}, arXiv:2405.15152.

\bibitem[{Hendrycks et~al.(2021)Hendrycks, Burns, Basart, Zou, Mazeika, Song, and Steinhardt}]{hendrycks2021measuringmassivemultitasklanguage}
Dan Hendrycks, Collin Burns, Steven Basart, Andy Zou, Mantas Mazeika, Dawn Song, and Jacob Steinhardt. 2021.
\newblock \href {https://arxiv.org/abs/2009.03300} {Measuring massive multitask language understanding}.
\newblock \emph{Preprint}, arXiv:2009.03300.

\bibitem[{Hine et~al.(2024)Hine, Novelli, Taddeo, and Floridi}]{Hine2024}
Emmie Hine, Claudio Novelli, Mariarosaria Taddeo, and Luciano Floridi. 2024.
\newblock \href {https://doi.org/10.1007/s11948-024-00500-5} {Supporting trustworthy ai through machine unlearning}.
\newblock \emph{Science and Engineering Ethics}, 30(5).

\bibitem[{Hong et~al.(2024)Hong, Yu, Yang, Ravfogel, and Geva}]{hong2024intrinsicevaluationunlearningusing}
Yihuai Hong, Lei Yu, Haiqin Yang, Shauli Ravfogel, and Mor Geva. 2024.
\newblock \href {https://arxiv.org/abs/2406.11614} {Intrinsic evaluation of unlearning using parametric knowledge traces}.
\newblock \emph{Preprint}, arXiv:2406.11614.

\bibitem[{Ji et~al.(2024)Ji, Liu, Zhang, Liu, Kompella, Liu, and Chang}]{ji2024reversingforgetretainobjectivesefficient}
Jiabao Ji, Yujian Liu, Yang Zhang, Gaowen Liu, Ramana~Rao Kompella, Sijia Liu, and Shiyu Chang. 2024.
\newblock \href {https://arxiv.org/abs/2406.08607} {Reversing the forget-retain objectives: An efficient llm unlearning framework from logit difference}.
\newblock \emph{Preprint}, arXiv:2406.08607.

\bibitem[{Jia et~al.(2024)Jia, Liu, Zhang, Ram, Baracaldo, and Liu}]{jia2024waglestrategicweightattribution}
Jinghan Jia, Jiancheng Liu, Yihua Zhang, Parikshit Ram, Nathalie Baracaldo, and Sijia Liu. 2024.
\newblock \href {https://arxiv.org/abs/2410.17509} {Wagle: Strategic weight attribution for effective and modular unlearning in large language models}.
\newblock \emph{Preprint}, arXiv:2410.17509.

\bibitem[{Liu et~al.(2024{\natexlab{a}})Liu, Yao, Jia, Casper, Baracaldo, Hase, Yao, Liu, Xu, Li, Varshney, Bansal, Koyejo, and Liu}]{liu2024rethinkingmachineunlearninglarge}
Sijia Liu, Yuanshun Yao, Jinghan Jia, Stephen Casper, Nathalie Baracaldo, Peter Hase, Yuguang Yao, Chris~Yuhao Liu, Xiaojun Xu, Hang Li, Kush~R. Varshney, Mohit Bansal, Sanmi Koyejo, and Yang Liu. 2024{\natexlab{a}}.
\newblock \href {https://arxiv.org/abs/2402.08787} {Rethinking machine unlearning for large language models}.
\newblock \emph{Preprint}, arXiv:2402.08787.

\bibitem[{Liu et~al.(2024{\natexlab{b}})Liu, Dou, Tan, Tian, and Jiang}]{liu-etal-2024-towards-safer}
Zheyuan Liu, Guangyao Dou, Zhaoxuan Tan, Yijun Tian, and Meng Jiang. 2024{\natexlab{b}}.
\newblock \href {https://doi.org/10.18653/v1/2024.findings-acl.107} {Towards safer large language models through machine unlearning}.
\newblock In \emph{Findings of the Association for Computational Linguistics: ACL 2024}, pages 1817--1829, Bangkok, Thailand. Association for Computational Linguistics.

\bibitem[{Neel and Chang(2024)}]{neel2024privacyissueslargelanguage}
Seth Neel and Peter Chang. 2024.
\newblock \href {https://arxiv.org/abs/2312.06717} {Privacy issues in large language models: A survey}.
\newblock \emph{Preprint}, arXiv:2312.06717.

\bibitem[{Pawelczyk et~al.(2023)Pawelczyk, Neel, and Lakkaraju}]{pawelczyk2023context}
Martin Pawelczyk, Seth Neel, and Himabindu Lakkaraju. 2023.
\newblock In-context unlearning: Language models as few shot unlearners.
\newblock \emph{arXiv preprint arXiv:2310.07579}.

\bibitem[{Pochinkov and Schoots(2024)}]{pochinkov2024dissectinglanguagemodelsmachine}
Nicholas Pochinkov and Nandi Schoots. 2024.
\newblock \href {https://arxiv.org/abs/2403.01267} {Dissecting language models: Machine unlearning via selective pruning}.
\newblock \emph{Preprint}, arXiv:2403.01267.

\bibitem[{Qu et~al.(2024)Qu, Ding, Sun, Thilakarathna, Zhu, and Niyato}]{qu2024frontierdataerasuremachine}
Youyang Qu, Ming Ding, Nan Sun, Kanchana Thilakarathna, Tianqing Zhu, and Dusit Niyato. 2024.
\newblock \href {https://arxiv.org/abs/2403.15779} {The frontier of data erasure: Machine unlearning for large language models}.
\newblock \emph{Preprint}, arXiv:2403.15779.

\bibitem[{Ramakrishna et~al.(2025)Ramakrishna, Wan, Jin, Chang, Bu, Vinzamuri, Cevher, Hong, and Gupta}]{ramakrishna2025lumellmunlearningmultitask}
Anil Ramakrishna, Yixin Wan, Xiaomeng Jin, Kai-Wei Chang, Zhiqi Bu, Bhanukiran Vinzamuri, Volkan Cevher, Mingyi Hong, and Rahul Gupta. 2025.
\newblock \href {https://arxiv.org/abs/2502.15097} {Lume: Llm unlearning with multitask evaluations}.
\newblock \emph{Preprint}, arXiv:2502.15097.

\bibitem[{Thaker et~al.(2024)Thaker, Maurya, Hu, Wu, and Smith}]{thaker2024guardrailbaselinesunlearningllms}
Pratiksha Thaker, Yash Maurya, Shengyuan Hu, Zhiwei~Steven Wu, and Virginia Smith. 2024.
\newblock \href {https://arxiv.org/abs/2403.03329} {Guardrail baselines for unlearning in llms}.
\newblock \emph{Preprint}, arXiv:2403.03329.

\bibitem[{Wu et~al.(2023)Wu, Li, Xu, Dong, Wu, Bian, and Xiong}]{wu-etal-2023-depn}
Xinwei Wu, Junzhuo Li, Minghui Xu, Weilong Dong, Shuangzhi Wu, Chao Bian, and Deyi Xiong. 2023.
\newblock \href {https://doi.org/10.18653/v1/2023.emnlp-main.174} {{DEPN}: Detecting and editing privacy neurons in pretrained language models}.
\newblock In \emph{Proceedings of the 2023 Conference on Empirical Methods in Natural Language Processing}, pages 2875--2886, Singapore. Association for Computational Linguistics.

\bibitem[{Xia et~al.(2024)Xia, Kim, Chen, Ye, Kundu, Hao, and Talati}]{xia2024understandingperformanceestimatingcost}
Yuchen Xia, Jiho Kim, Yuhan Chen, Haojie Ye, Souvik Kundu, Cong Hao, and Nishil Talati. 2024.
\newblock \href {https://arxiv.org/abs/2408.04693} {Understanding the performance and estimating the cost of llm fine-tuning}.
\newblock \emph{Preprint}, arXiv:2408.04693.

\bibitem[{Yao et~al.(2024)Yao, Xu, and Liu}]{yao2024largelanguagemodelunlearning}
Yuanshun Yao, Xiaojun Xu, and Yang Liu. 2024.
\newblock \href {https://arxiv.org/abs/2310.10683} {Large language model unlearning}.
\newblock \emph{Preprint}, arXiv:2310.10683.

\bibitem[{Yu et~al.(2023)Yu, Jeoung, Kasi, Yu, and Ji}]{yu-etal-2023-unlearning}
Charles Yu, Sullam Jeoung, Anish Kasi, Pengfei Yu, and Heng Ji. 2023.
\newblock \href {https://doi.org/10.18653/v1/2023.findings-acl.375} {Unlearning bias in language models by partitioning gradients}.
\newblock In \emph{Findings of the Association for Computational Linguistics: ACL 2023}, pages 6032--6048, Toronto, Canada. Association for Computational Linguistics.

\bibitem[{Yuan et~al.(2024)Yuan, Jin, Cao, Chen, Liu, and Zhao}]{yuan2024robustknowledgeunlearningadversarial}
Hongbang Yuan, Zhuoran Jin, Pengfei Cao, Yubo Chen, Kang Liu, and Jun Zhao. 2024.
\newblock \href {https://arxiv.org/abs/2408.10682} {Towards robust knowledge unlearning: An adversarial framework for assessing and improving unlearning robustness in large language models}.
\newblock \emph{Preprint}, arXiv:2408.10682.

\bibitem[{Zhang et~al.(2023)Zhang, Chen, Liu, and He}]{zhang2023composingparameterefficientmodulesarithmetic}
Jinghan Zhang, Shiqi Chen, Junteng Liu, and Junxian He. 2023.
\newblock \href {https://arxiv.org/abs/2306.14870} {Composing parameter-efficient modules with arithmetic operations}.
\newblock \emph{Preprint}, arXiv:2306.14870.

\bibitem[{Zhang et~al.(2025)Zhang, Li, Qian, Jiang, and Chen}]{ZHANG2025100301}
Ran Zhang, Hong-Wei Li, Xin-Yuan Qian, Wen-Bo Jiang, and Han-Xiao Chen. 2025.
\newblock \href {https://doi.org/10.1016/j.jnlest.2025.100301} {On large language models safety, security, and privacy: A survey}.
\newblock \emph{Journal of Electronic Science and Technology}, page 100301.

\bibitem[{Zhang et~al.(2024)Zhang, Lin, Bai, and Mei}]{zhang2024negative}
Ruiqi Zhang, Licong Lin, Yu~Bai, and Song Mei. 2024.
\newblock Negative preference optimization: From catastrophic collapse to effective unlearning.
\newblock \emph{arXiv preprint arXiv:2404.05868}.

\bibitem[{Zhao et~al.(2024)Zhao, Wu, Nguyen, Jia, Feng, and Tuan}]{zhao2024unlearningbackdoorattacksllms}
Shuai Zhao, Xiaobao Wu, Cong-Duy Nguyen, Meihuizi Jia, Yichao Feng, and Luu~Anh Tuan. 2024.
\newblock \href {https://arxiv.org/abs/2410.14425} {Unlearning backdoor attacks for llms with weak-to-strong knowledge distillation}.
\newblock \emph{Preprint}, arXiv:2410.14425.

\end{thebibliography}
\appendix

\section{Other experiments conducted}
\label{sec:B}
\begin{itemize}
 \item \textbf{Prompt routing}\\
If there are any distinguishable patterns between the samples of forget and retain sets, they can be used to train the target model respond in a certain way when a sample is identified to be from the forget set distribution and vice versa. These patterns might be identifiable through sample clustering by grouping similar ones. Additionally, as the labels are available for the forget and retain samples, a binary classifier can be trained to classify the samples. If the samples are effectively classifiable, it therefore, can be used to classify the input, and accordingly make the target model respond.
\begin{figure*}[ht!]
     \centering
     \begin{subfigure}[ht]{0.42\textwidth}
         \centering
         \includegraphics[width=\textwidth]{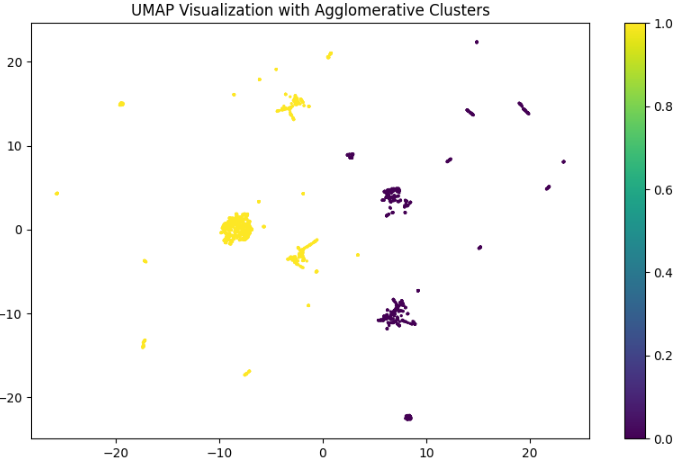}
         \caption{Agglomerative cluster distribution}
     \end{subfigure}
     \hfill
     \begin{subfigure}[ht]{0.42\textwidth}
         \centering
         \includegraphics[width=\textwidth]{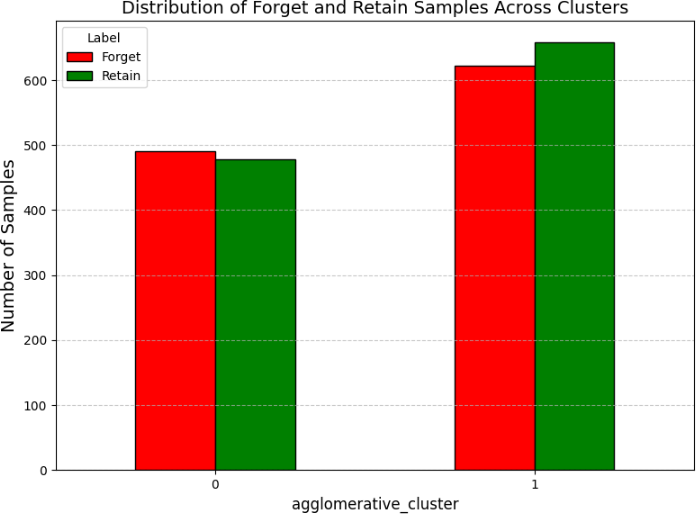}
         \caption{Forget and Retain samples distribution across agglomerative clusters}
     \end{subfigure}

     \begin{subfigure}[ht]{0.42\textwidth}
         \centering
         \includegraphics[width=\textwidth]{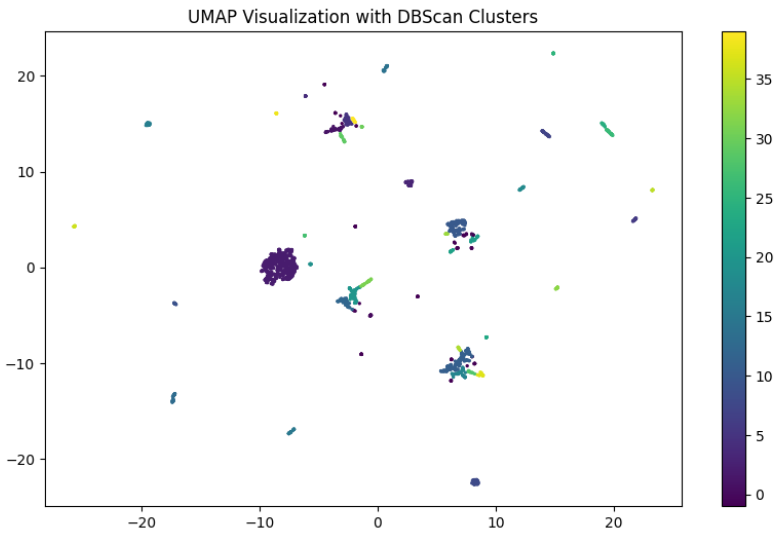}
         \caption{DBSCAN cluster distribution}
     \end{subfigure}
     \hfill
     \begin{subfigure}[ht]{0.42\textwidth}
         \centering
         \includegraphics[width=\textwidth]{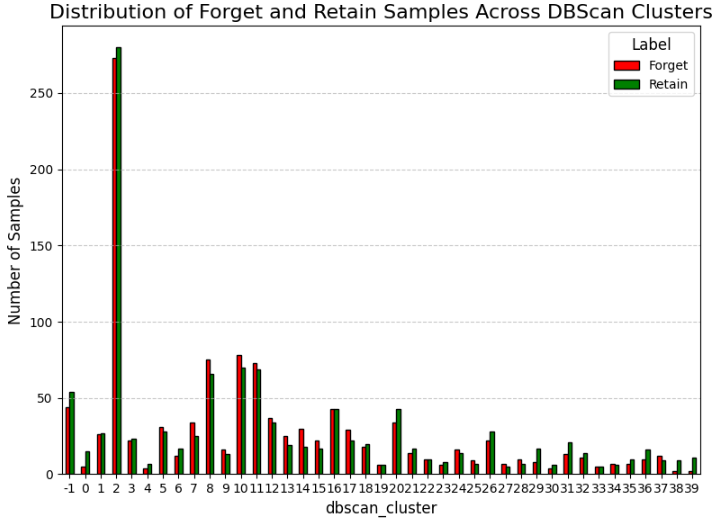}
         \caption{Forget and Retain samples distribution across DBSCAN clusters}
     \end{subfigure}
     \hfill
             \caption{Clusters visualization, and distribution of forget and retain samples across respective clusters}
        \label{fig:cluster}
\end{figure*}

\begin{table}[bp]
  \centering
  \begin{tabular}{lll}
    \toprule
    \textbf{Split} & \textbf{Train} & \textbf{Test} \\\midrule
    Forget\_train     & 890 & 222           \\
    Retain\_train     & 909 & 227           \\\bottomrule
  \end{tabular}
  \caption{Distribution of train and test set samples used for classifier training}
  \label{tab:denomination}
\end{table}

\begin{itemize}
\item \textbf{Clustering}\\
Two distinct clustering algorithms were used: Agglomerative clustering, Density-Based Spatial Clustering of Applications with Noise (DBSCAN). In agglomerative clustering, which is a hierarchical approach, it is required to specify the desired number of clusters, which was set to two here, denoting the forget and retain sets. Unlike agglomerative, DBSCAN is a density-based clustering approach, where $epsilon$ was set to 0.3, and $minimum\_samples$ was set to 10.
\item \textbf{Classification}\\
Six machine learning algorithms and an ensemble soft voting classifier were trained on the combined forget and retain datasets, with an 80:20 train-test split. The denomination of samples in the train and the test sets are reported in Table \ref{tab:denomination}. 
\end{itemize}
\begin{figure*}[t]
  \includegraphics[width=\textwidth]{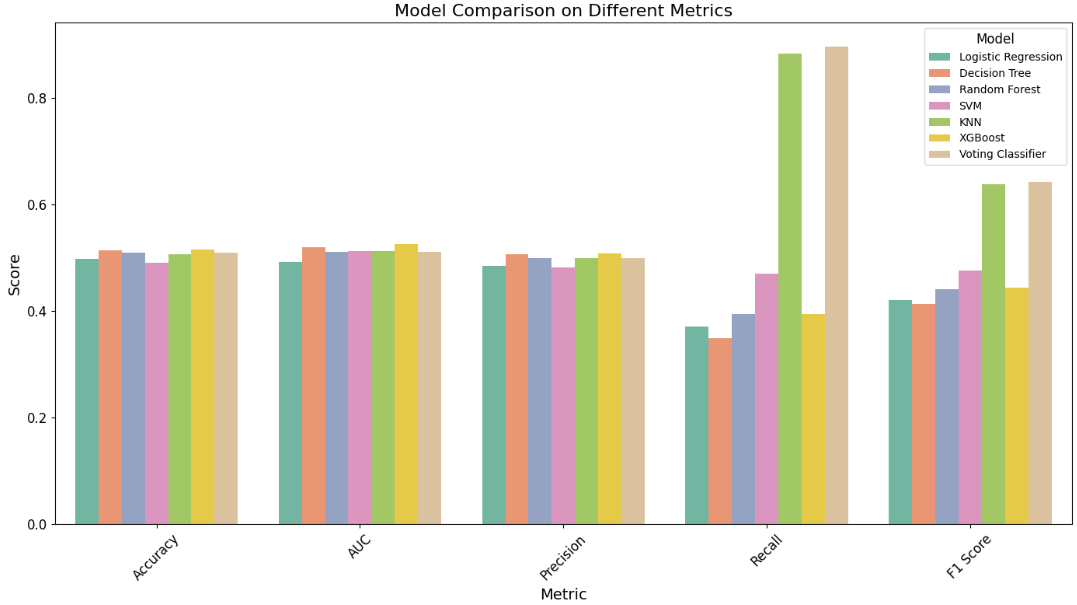}
  \caption{Binary classifier's performance across metrics}
  \label{fig:classify}
\end{figure*}

\textbf{Observations:} The features used to represent the samples result in significant overlap in the feature space, failing to provide sufficient separation between the two disjoint classes. This is evident in both clustering and classification results. Clustering, using both predefined and non-predefined groups, showed that each resulting cluster contained a proportionate number of samples from both classes. Furthermore, classification demonstrated that no tested classifier achieved significant performance metrics. These results are illustrated in Figures \ref{fig:cluster} and \ref{fig:classify}. Although these results were obtained using the OLMo-1B-0724-hf tokenizer (the default for the 1B model), the observations remain consistent across other tested tokenizers: deberta-v3-large and all\_Mini\_LM.
\vspace{-1mm}
\item \textbf{Logits difference:}
Drawing from \cite{ji2024reversingforgetretainobjectivesefficient}, to use an assistant model which is trained to remember the forget set, whose logits when subtracted from that of the target model will result in effective unlearning of the forget set for the queries inferred upon, we experimented with the following directions. For all the experiments, a temperature of zero was set, and a scaling factor of 0.2 was used for subtracting logits.
\begin{table}[!ht]
\centering
\begin{tabular}{ll}
\toprule
\textbf{Layer}     & \textbf{Modify ratio} \\ \toprule
self\_attn.q\_proj & 0                     \\ \midrule
self\_attn.k\_proj & 0.00001               \\ \midrule
self\_attn.v\_proj & 0.0001                \\ \midrule
self\_attn.o\_proj & 0.01                  \\ \midrule
mlp.gate\_proj     & 0.03                  \\ \midrule
mlp.up\_proj       & 0                     \\ \midrule
mlp.down\_proj     & 0.07                  \\ \midrule
First 12 layers    & 0 (freezed)           \\ \bottomrule
\end{tabular}
\caption{Layer-wise perturbation configuration}
\label{truncation_table}
\vspace{-0.952em}
\end{table}
\begin{itemize}
    \item \textbf{Reinitializing the weights of pretrained model, and tuning on the forget set:} An assistant model was prepared with a copy of the target model's configuration (1B). This model was initialized with Xavier initialization, and was trained on the forget set. As the forget set is very small with very limited samples to train a model, the assistant model resulted in generating garbage characters, sometimes, repeated words without a complete meaning. Therefore, this did not result in effective unlearning, as the assistant model could not pick up the forget set due to its small quantity. This experiment emphasizes that, in cases, the size of the forget set plays an important role to understand the information to be forgotten effectively, and availability or provision of only limited forget samples could lead to ineffective unlearning of the target model. 
    \item \textbf{Using another domain-irrelevant language model as the assistant:} Based on the above observation, a hypothesis was formulated as -- instead of using an assistant model with no prior knowledge, if it has certain level of language understanding, it might be able to pick up the forget set despite its limited quantity. Thus, fintech-chatbot-t5\footnote{\url{https://huggingface.co/cuneytkaya/fintech-chatbot-t5}}, a small domain-irrelevant model was considered for the assistant model. This model is based on T5-small architecture and was trained on the retail banking chatbot dataset\footnote{\url{https://huggingface.co/datasets/bitext/Bitext-retail-banking-llm-chatbot-training-dataset/}} for only 3 epochs. We have finetuned this model on the forget dataset. However, it was observed that the logits difference did not give any meaningful output when decoded. Apparently, due to different tokenizers used by the assistant and the target models, the encoded vectors were not aligned to be subtracted. 
    \item \textbf{Knowledge truncation in the pretrained model being used as the assistant:} Considering the discussed results, a copy of the target model was used as the assistant model, truncating its knowledge, such that it preserves the language understanding capabilities, and other general abilities, but not the specific subject matter expertise, or any particular details. The target model has 16 layers in total, and each layer has 7 components: 4 for self-attention and 3 for feed-forward network. A brute-force approach was followed to identify the best (better) combination of layers that are required to be retained as is, and the layers that are required to be perturbed. The perturbation method followed was to add a factor of noise determined by $torch.randn\_like(param.data) * modify\_ratio$ where $param.data$ is the corresponding parametric value for a parameter in a layer, and the combination of $modify\_ratio$s that worked decently are reported in Table \ref{truncation_table}. Although this experiment worked fairly based on the limited combinations tested with, the responses still required significant refinement, demanding rigorous testing. 
\end{itemize}
\end{itemize}


\section{Training configuration and detailed results}
\label{sec:A}
\begin{table}[H]

\centering
\hspace*{1.5cm}
\footnotesize
\setlength{\tabcolsep}{2pt} 
\begin{tabular}{p{4.5cm}p{8cm}}
\toprule
\textbf{1B model training configs} & \textbf{7B model training configs} \\ \midrule
\begin{itemize}[noitemsep,nolistsep,leftmargin=*] 
    \item batch\_size = 8
    \item AdamW optimizer
    \item Linear scheduler
    \item num\_warmup\_steps = 3
    \item gradient\_clip's max\_norm = 1
    \item tokenization: 
        \begin{itemize}[noitemsep,nolistsep,leftmargin=*]
            \item max\_length = 512
            \item truncation = True
            \item padding = 'max\_length'
        \end{itemize}
    \item For GA: Loss = -outputs.loss 
\end{itemize} 
& 
\begin{itemize}[noitemsep,nolistsep,leftmargin=*]
    \item Quantization (BitsAndBytesConfig): 
        \begin{itemize}[noitemsep,nolistsep,leftmargin=*]
            \item load\_in\_4bit = True
            \item bnb\_4bit\_quant\_type = "nf4"
            \item bnb\_4bit\_compute\_dtype = "float16"
        \end{itemize}
    \item PEFT (LORA) config:
        \begin{itemize}[noitemsep,nolistsep,leftmargin=*]
            \item lora\_alpha = 16
            \item lora\_dropout = 0.1
            \item r = 64
            \item target\_modules = \{'q\_proj', 'k\_proj', 'v\_proj',
                                   'o\_proj', 'gate\_proj', 'upd\_proj', 'down\_proj'\} 
            \item bias = "none"
            \item task\_type = "CAUSAL\_LM"
        \end{itemize}
    \item Training config:
        \begin{itemize}[noitemsep,nolistsep,leftmargin=*]
            \item per\_device\_train\_batch\_size = 4
            \item SFTTrainer
            \item formatting\_func returns list of input, output pairs
            \item For GA: Loss = -outputs.loss
        \end{itemize}
\end{itemize} \\ \bottomrule
\end{tabular}
\captionsetup{width=1.9\linewidth,justification=raggedleft,
    singlelinecheck=false} 
\caption{Training configurations of 1B and 7B models}

\label{training_config_table}
\end{table}

\begin{table*}[ht]
\centering
\setlength{\tabcolsep}{2pt} 
\footnotesize 
\begin{tabular}{p{2.5cm}p{1.6cm}p{1.1cm}p{3cm}p{1.3cm}p{5cm}} 
\toprule
\textbf{Method} & \textbf{Aggregate} & \textbf{Task Agg.} & \textbf{MIA score} & \textbf{MMLU Avg Acc.} & \textbf{Configuration} \\ \midrule
\begin{tabular}[c]{@{}l@{}}Gradient\\ 
Ascent (GA)\end{tabular} & 
    \begin{tabular}[c]{@{}l@{}}0.181\\ 0.345\end{tabular} & 
    \begin{tabular}[c]{@{}l@{}}0.222\\ 0\end{tabular} & 
    \begin{tabular}[c]{@{}l@{}}0.049\\ 0.807\end{tabular} & 
    \begin{tabular}[c]{@{}l@{}}0.271\\ 0.229\end{tabular} & 
    \begin{tabular}[c]{@{}l@{}}LR=2e-7, WD=2e-6, E = 6; \\
    
    LR = 2e-5, WD = 2e-4, E = 3\end{tabular}\\ \midrule
Controlled GA & 0.370 & 0 & 0.855 & 0.255 & LR=2e-5, E=10, No WD, alpha=0.1 \\ \midrule
Gradient 

Descent (GD) & 
    0.231
    
    0.410 & 
    0
    
    0 & 
    0.417
    
    0.982 & 
    0.275
    
    0.247 & LR = 2e-6; WD = 2e-5; E = 6; 
    
    LR=2e-5; WD=2e-4; LR\_Scheduler = Cosine schedule; E=20\\ \midrule
Gradient 

Difference & 
    \begin{tabular}[c]{@{}l@{}}0.323\\ 0.325\\ 0.290\\ 0.316\\ 0.302\\ 0.331\\ 0.336\\ 0.345\\ 0.360\end{tabular} & 
    \begin{tabular}[c]{@{}l@{}}0\\ 0\\ 0\\ 0\\ 0\\ 0\\ 0\\ 0\\ 0\end{tabular} & 
    \begin{tabular}[c]{@{}l@{}}0.701 (6 GD epochs)\\ 0.711 (9 GD epochs)\\ 0.621 (12 GD epochs)\\ 0.687 (15 GD epochs)\\ 0.670 (18 GD epochs)\\ 0.742 (21 GD epochs)\\ 0.753 (24 GD epochs)\\ 0.800 (27 GD epochs)\\ 0.825 (30 GD epochs)\end{tabular} & 
    \begin{tabular}[c]{@{}l@{}}0.269\\ 0.263\\ 0.248\\ 0.260\\ 0.237\\ 0.251\\ 0.254\\ 0.235\\ 0.255\end{tabular} & 
    GA(LR=2e-7, WD=2e-6, E = 6) -> GD (First 6 GD epochs: LR = 2e-7, WD = 2e-6;
    After that: LR = 2e-5, WD = 2e-4) \\ \midrule
KL Minimization & 0.174 & 0.219 & 0.032 & 0.272 & LR=2e-7, WD=2e-6, E = 6 \\ \midrule
Xavier init (1B) & 0.402 & 0 & 0.944 & 0.261 & Original model weights are erased and initialized with Xavier initialization method\\ \midrule
Original 

model (1B) & 0.0913 & 0 & 0 & 0.274 & - \\ \bottomrule
\end{tabular}
\caption{Performance of 1B model -- A comprehensive view
(LR: Learning rate; WD: Weight decay; E: Epoch)}
\label{1B_full_table}
\vspace{1em}
\end{table*}

\begin{table*}[!ht]
\centering
\setlength{\tabcolsep}{4pt} 
\footnotesize
\begin{tabular}{p{2.5cm}p{1.7cm}p{1cm}p{1.1cm}p{1.3cm}p{6cm}}  
\toprule
\textbf{Method} & \textbf{Aggregate} & \textbf{Task Agg.} & \textbf{MIA score} & \textbf{MMLU Avg Acc.} & \textbf{Configuration} \\ \midrule

Gradient 

Ascent (GA) & 0.383 & 0 & 0.865 & 0.284 & LR= 2e-6, E=10 \\ \midrule

Gradient 

Descent (GD) & 0.170 & 0.005 & 0 & 0.504 & LR=2e-5, E=20 \\ \midrule

Gradient 

Difference 

(GDf) & 
\makecell[l]{\\0.170 \\ 0.169 \\ 0.168 \\ 0.171 \\ 0.170} & 
\makecell[l]{\\0 \\ 0 \\ 0 \\ 0 \\ 0} & 
\makecell[l]{\\0 \\ 0 \\ 0 \\ 0 \\ 0} & 
\makecell[l]{\\0.504 \\ 0.502 \\ 0.505 \\ 0.512 \\ 0.511} & 
\makecell[l]{GA: LR= 2e-6, E=10 \\ GD: LR= 2e-4, E=25 \\ GD: LR= 2e-4, E=5 \\ GD: LR= 2e-4, E=3 \\ GD: LR= 2e-6, E=3 \\ GD: LR= 2e-6, E=1} \\ \midrule

GDf -> GA & 
\makecell[l]{0.377 \\ \\0.442 \\ \\0.447 \\ \\0.170} & 
\makecell[l]{0 \\ \\0 \\ \\0 \\ \\0} & 
\makecell[l]{0.67 \\ \\0.982 \\ \\0.998 \\ \\0} & 
\makecell[l]{0.461 \\ \\0.345 \\ \\0.343 \\ \\0.511} & 
\makecell[l]{GA (LR= 2e-6, E=10) -> GD (E=3: LR= 2e-6) \\-> GA (E=1: LR= 2e-5) \\ GA (LR= 1e-4, E=3) -> GD (E=3: LR= 2e-6) \\-> GA (E=1: LR= 2e-5) \\ GA (LR= 1e-5, E=3) -> GD (E=3: LR= 2e-6) \\-> GA (E=1: LR= 2e-5) \\ GA (LR= 1e-5, E=3) -> GD (E=3: LR= 2e-6) \\-> GA (E=1: LR= 2e-6)} \\ \midrule

GDf -> GDf & 
\makecell[l]{0.171 \\ \\0.170} & 
\makecell[l]{0 \\ \\0} & 
\makecell[l]{0 \\ \\0} & 
\makecell[l]{0.513 \\ \\0.511} & 
\makecell[l]{GA (LR= 1e-4, E=3) -> GD (LR= 2e-6, E=3) \\-> GA (LR= 2e-5, E=1) -> GD (LR= 2e-6, E=1) \\ GA (LR= 1e-4, E=3) -> GD (LR= 2e-6, E=3) \\-> GA (LR= 2e-5, E=1) -> GD (LR= 2e-8, E=1)} \\ \midrule

Xavier init (7B) & 0.397 & 0 & 0.936 & 0.255 & Original model weights are erased and initialized with Xavier initialization method \\ \midrule

Original model (7B) & 0.170 & 0 & 0 & 0.512 & - \\ \midrule

GD -> GA & 
\makecell[l]{0.170 \\ 0.365} & 
\makecell[l]{0 \\ 0} & 
\makecell[l]{0 \\ 0.847} & 
\makecell[l]{0.509 \\ 0.247} & 
\makecell[l]{GD: LR=2e-5, E=20; GA: LR= 2e-6, E=3 \\ GD: LR=2e-5, E=20; GA: LR= 2e-4, E=3} \\ \bottomrule

\end{tabular}
\caption{Performance of 7B model -- A comprehensive view (LR: Learning rate; E: Epoch)}
\label{7B_full_table}
\end{table*}


\end{document}